\pdfoutput=1

\documentclass[11pt]{article}

\usepackage[]{acl}

\usepackage{times}
\usepackage{latexsym}
\usepackage{soul}
\usepackage{url}
\usepackage[utf8]{inputenc}
\usepackage{graphicx}
\usepackage{amsmath}
\usepackage{amsthm}
\usepackage{booktabs}
\usepackage{algorithm}
\usepackage{algorithmic}
\usepackage{multirow}
\usepackage{amssymb}
\usepackage{bbding}
\usepackage{mathrsfs}
\usepackage{longtable}
\usepackage[T1]{fontenc}
\usepackage{enumitem}
\usepackage{float}

\usepackage[utf8]{inputenc}
\usepackage{makecell}
\usepackage{microtype}

\title{Evaluating ChatGPT’s Information Extraction Capabilities: An Assessment of Performance, Explainability, Calibration, and Faithfulness}

\author{
        Bo Li$^{1,2}$, Gexiang Fang$^{1,2}$, Yang Yang$^{1,2}$, Quansen Wang$^{3}$, \\
        \textbf{Wei Ye}$^{1}$, \textbf{Wen Zhao}$^{1}$, and \textbf{Shikun Zhang}$^{1}$\\
	$^1$National Engineering Research Center for Software Engineering, Peking University\\
	$^2$School of Software and Microelectronics, Peking University\\
	$^3$Boston University\\
        \{deepblue.lb, fanggx, yangy\}@stu.pku.edu.cn,  
        quansenw@bu.edu \\
	\{wye, zhaowen, zhangsk\}@pku.edu.cn
}

\begin{document}
\maketitle
\begin{abstract}

The capability of Large Language Models (LLMs) like ChatGPT to comprehend user intent and provide reasonable responses has made them extremely popular lately. In this paper, we focus on assessing the overall ability of ChatGPT using 7 fine-grained information extraction (IE) tasks. Specially, we present the systematically analysis by measuring ChatGPT's performance, explainability, calibration, and faithfulness, and resulting in 15 keys from either the ChatGPT or domain experts. Our findings reveal that ChatGPT’s performance in Standard-IE setting is poor, but it surprisingly exhibits excellent performance in the OpenIE setting, as evidenced by human evaluation. In addition, our research indicates that ChatGPT provides high-quality and trustworthy explanations for its decisions. However, there is an issue of ChatGPT being overconfident in its predictions, which resulting in low calibration. Furthermore, ChatGPT demonstrates a high level of faithfulness to the original text in the majority of cases. We manually annotate and release the test sets of 7 fine-grained IE tasks contains 14 datasets to further promote the research. The datasets and code are available at \href{https://github.com/pkuserc/ChatGPT_for_IE}{this url}.~\footnote{https://github.com/pkuserc/ChatGPT\_for\_IE}

\end{abstract}

\section{Introduction}

Large Language Models (LLMs) (e.g., GPT-3~\cite{DBLP:conf/nips/BrownMRSKDNSSAA20}, LaMDA~\cite{DBLP:journals/corr/abs-2201-08239} and PaLM~\cite{DBLP:journals/corr/abs-2204-02311}, etc.) have greatly promoted the development of the Natural Language Processing (NLP) community. With a proper instruction (often the task definition)~\cite{DBLP:journals/corr/abs-2203-02155,DBLP:journals/corr/abs-2205-11916,chung2022scaling,DBLP:conf/emnlp/WangMAKMNADASPK22} and the chain-of-thought (CoT) prompting~\cite{DBLP:journals/corr/abs-2201-11903}, LLMs achieve surprisingly good performances when dealing with unseen tasks. 

ChatGPT~\footnote{https://chat.openai.com/} is currently the most popular LLM, known for its impressive ability to understand user intent and generate human-like responses. ChatGPT is trained on the GPT family~\cite{DBLP:conf/nips/BrownMRSKDNSSAA20,DBLP:conf/emnlp/ArtetxeBGMOSLDI22,DBLP:journals/corr/abs-2203-02155} using reinforcement learning from human feedback
(RLHF)~\cite{DBLP:conf/nips/ChristianoLBMLA17} and high-quality conversational-style datasets. Apart from its surprising dialogue ability, ChatGPT has many other aspects that attract researchers to explore. Some researchers have delved into the potential impacts of ChatGPT on human life~\cite{DBLP:journals/corr/abs-2212-05856,DBLP:journals/corr/abs-2301-12867,DBLP:journals/corr/abs-2212-09292,DBLP:journals/corr/abs-2302-04536}. Other researchers are interested in exploring the capabilities of ChatGPT for various NLP tasks~\cite{DBLP:journals/corr/abs-2212-14548,qin2023chatgpt,DBLP:journals/corr/abs-2301-13852,DBLP:journals/corr/abs-2301-07597}. The capabilities of ChatGPT have been preliminarily explored through the above research and valuable conclusions have been drawn.

Given ChatGPT is a closed model that does not provide information about its training details, and any response from the model encodes an opinion. The response can significantly impact the user’s experience and shape their beliefs going forward~\cite{aiyappa2023can,santurkar2023whose,deshpande2023toxicity,DBLP:journals/corr/abs-2302-07736}. Consequently, evaluating ChatGPT should involve not only assessing its ability to achieve high performance but also measuring the reliability of the answers it provides. To help users better understand the overall quality of ChatGPT’s responses and enable systematic measurement of its capabilities, we design the following four metric dimensions: The first dimension we consider is \textbf{Performance}, which reflects ChatGPT's overall performance on various IE tasks from multiple perspectives. The second metric dimension, \textbf{Explainability}~\cite{DBLP:conf/acl/RajaniMXS19,DBLP:conf/acl/AghajanyanGZ20,DBLP:journals/csur/ZiniA23}, evaluates whether ChatGPT could give a justified reason for its prediction, thereby providing insights into ChatGPT's decision-making process. The third one is \textbf{Calibration}~\cite{DBLP:conf/icml/GuoPSW17,DBLP:conf/nips/KumarLM19,DBLP:conf/nips/ThulasidasanCBB19,DBLP:conf/nips/MindererDRHZHTL21}, which measures the predictive uncertainty of a model, and we use this metric to assess if ChatGPT is overconfidence on its prediction. The last dimension is \textbf{Faithfulness}~\cite{DBLP:conf/acl/MaynezNBM20,koto2022can,DBLP:journals/corr/abs-2208-14271,DBLP:journals/corr/abs-2301-00303}, it is frequently employed in the summarization task to determine whether the summary accurately reflects the input. In our research, we adopt faithfulness as a measure of whether the explanations given by ChatGPT are truthful to the input, or if they are spurious. In summary, according to the above four dimensions, we collect 15 keys from either the ChatGPT or domain experts for the evaluation  (\S~\ref{sec:data_collection}).

In this research, we aim to perform a comprehensive study and detailed analysis of ChatGPT's capabilities through various information extraction (IE) tasks. IE involves heterogeneous structure extraction, factual knowledge usage, and diversified targets\cite{DBLP:conf/emnlp/YamadaASTM20,DBLP:conf/iclr/PaoliniAKMAASXS21,lu-etal-2022-unified}, making it an ideal scenario for evaluating ChatGPT's capabilities. Overall, we conduct our experiments and analysis based on 14 datasets belonging to 7 fine-grained IE tasks(\S~\ref{sec:main}). Additionally, we assess the explainability, calibration, and faithfulness of ChatGPT's responses through both \textit{self-check} and \textit{human-check} (\S~\ref{sec:analysis}). To sum up, our main contributions are summarized as follows:

\begin{itemize}
    \item To assess the overall ability of ChatGPT, we employ a comprehensive and systematic evaluation from four dimensions: 1) performance, 2) explainability, 3) calibration, and 4) faithfulness. We then collected 15 keys belonging to above dimensions from either the ChatGPT or domain experts for the research. All the manually annotated datasets and code are made public available for future research.

    \item We comprehensively evaluate the overall performance of ChatGPT on various tasks in both Standard-IE and OpenIE settings and compare it with other popular models. Our research indicates that ChatGPT’s performance is not satisfactory in the Standard-IE setting. However, we show that it provides surprisingly good results in the OpenIE setting, as confirmed by human evaluation. Furthermore, we also discover that ChatGPT provides high-quality and trustworthy explanations for its decisions. Although, it displays overconfidence in its predictions, leading to low calibration. Besides, ChatGPT is largely faithful to the original text in most cases. 
\end{itemize}

\section{Related Work}

\subsection{Large Language Models}

Large Language Models (LLMs) typically contain more than a hundred billion parameters, such as GPT-3~\cite{DBLP:conf/nips/BrownMRSKDNSSAA20}, Gopher~\cite{DBLP:journals/corr/abs-2112-11446}, LaMDA~\cite{DBLP:journals/corr/abs-2201-08239}, Megatron-turing-NLG~\cite{DBLP:journals/corr/abs-2201-11990}, and PaLM~\cite{DBLP:journals/corr/abs-2204-02311}, among others. Scaling up the model size brings impressive abilities on few-shot and zero-shot learning scenarios, such as producing reasonable results with very few samples or task descriptions \cite{DBLP:conf/nips/BrownMRSKDNSSAA20,DBLP:journals/corr/abs-2204-02311}. Moreover, scaling up the model size unlocks emergent abilities that were not observed in smaller models, enabling LLMs to exhibit strong generalizability on unseen tasks~\cite{DBLP:journals/corr/abs-2206-07682,fu2022does,DBLP:journals/corr/abs-2301-06627}.


With its impressive ability to understand user intent and generate human-like responses, ChatGPT has become the most popular language model currently. It is trained on the GPT family~\cite{DBLP:conf/nips/BrownMRSKDNSSAA20,DBLP:conf/emnlp/ArtetxeBGMOSLDI22,DBLP:journals/corr/abs-2203-02155} and high-quality conversational-style datasets using reinforcement learning from human feedback (RLHF)~\cite{DBLP:conf/nips/ChristianoLBMLA17}. 

Along with its dialogue ability, ChatGPT has other aspects that researchers are exploring. Some researchers show potential impacts of ChatGPT on human life, such as ethical risks~\cite{DBLP:journals/corr/abs-2212-05856,DBLP:journals/corr/abs-2301-12867,DBLP:journals/corr/abs-2301-07098}, the education sector~\cite{DBLP:journals/corr/abs-2212-09292,DBLP:journals/corr/abs-2302-04536,kortemeyer2023could} and the medical scenario~\cite{DBLP:journals/corr/abs-2301-13819,DBLP:journals/corr/abs-2301-10035,DBLP:journals/corr/abs-2212-14882}. Additionally, some researchers are keen to examine the potential of ChatGPT in addressing various natural language processing tasks. For instance, some works have examined ChatGPT's performance in stance detection~\cite{DBLP:journals/corr/abs-2212-14548}, linguistic and sentiment analysis~\cite{susnjak2023applying,ortega2023linguistic}, general NLP tasks~\cite{qin2023chatgpt,bian2023chatgpt,zhong2023can,DBLP:journals/corr/abs-2302-12095,DBLP:journals/corr/abs-2304-08085}, and machine translation~\cite{DBLP:journals/corr/abs-2301-08745}. \cite{DBLP:journals/corr/abs-2301-13867} explores the mathematical capabilities of ChatGPT, while \cite{DBLP:journals/corr/abs-2302-04023} proposes an evaluation of ChatGPT on reasoning and other aspects. Additionally, \cite{DBLP:journals/corr/abs-2301-13852,DBLP:journals/corr/abs-2301-07597} investigate the differences between human-written and ChatGPT-generated.

\subsection{Information Extraction}

Information Extraction (IE) is a long-standing research topic that aims to extract structured factual information from unstructured texts~\cite{DBLP:conf/anlp/AndersenHWHSN92,crowe-1995-constraint,chieu-etal-2003-closing,wu-weld-2010-open,khot-etal-2017-answering,lu-etal-2022-unified}. Typically, IE involves a wide range of tasks, such as named entity recognition(NER)~\cite{DBLP:conf/acl/GregoricBC18,DBLP:conf/acl/MartinsMM19,li-etal-2020-unified,DBLP:conf/acl/DasKPZ22}, entity typing(ET)~\cite{DBLP:conf/acl/LevyZCC18,DBLP:conf/acl/DaiSW20,DBLP:conf/acl/PangZZW22,DBLP:conf/acl/0019CJLZ0X22}, relation extraction(RE)~\cite{DBLP:conf/acl/LiYSLYCZL19,DBLP:conf/acl/FuLM19,DBLP:conf/acl/BianHHZZ21,DBLP:conf/acl/YeL0S22}, relation classification(RC)~\cite{zeng2015distant,DBLP:conf/acl/YeLXSCZ19,DBLP:journals/corr/abs-2102-01373,DBLP:journals/corr/abs-2212-14266}, event detection(ED)~\cite{DBLP:conf/acl/VeysehLDN20,DBLP:conf/acl/LouLDZC20,DBLP:conf/acl/LiuCX22,DBLP:conf/emnlp/ZhaoJ0GC22}, event argument extraction(EAE)~\cite{DBLP:conf/emnlp/ZhangSH22,DBLP:conf/emnlp/DuJ22,DBLP:conf/acl/MaW0LCWS22}, and event extraction(EE)~\cite{DBLP:conf/emnlp/WaddenWLH19,DBLP:conf/emnlp/DuC20,DBLP:conf/acl/LiuHSW22,DBLP:conf/naacl/HsuHBMNCP22}, among others. These tasks automatically generate structured factual outputs related to entity, and relation, and event, and greatly boost the development of NLP community.


\section{ChatGPT for Information Extraction}\label{sec:data_collection}
In this section, we first briefly introduce 7 fine-grained IE tasks, then we present how to collect 15 keys from the ChatGPT and domain experts.

\subsection{Information Extraction}

IE involves a wide range of tasks which need to extract structured factual information from unstructured texts, such as entity, and relation, and event. In this research, we conduct our analysis on the following 7 fine-grained IE tasks:\footnote{We introduce these tasks briefly due to the space limitation. Please refer to the task-specific papers for more details.} 1) \textbf{Entity Typing(ET)}~\cite{DBLP:conf/acl/LevyZCC18,DBLP:conf/acl/DaiSW20,DBLP:conf/acl/PangZZW22,DBLP:conf/acl/0019CJLZ0X22} aims to classify the type of a target entity under a given input; 2) \textbf{Named Entity Recognition(NER)}~\cite{DBLP:conf/acl/GregoricBC18,DBLP:conf/acl/MartinsMM19,li-etal-2020-unified,DBLP:conf/acl/DasKPZ22} aims to first identify the candidate entities, and then classify their types; 3) \textbf{Relation Classification(RC)}~\cite{zeng2015distant,DBLP:conf/acl/YeLXSCZ19,DBLP:journals/corr/abs-2102-01373,DBLP:journals/corr/abs-2212-14266} requires to classify the relation between two target entities; 4) \textbf{Relation Extraction(RE)}~\cite{DBLP:conf/acl/LiYSLYCZL19,DBLP:conf/acl/FuLM19,DBLP:conf/acl/BianHHZZ21,DBLP:conf/acl/YeL0S22} is a task to identify the target entities and the relation jointly; 5) \textbf{Event Detection(ED)}~\cite{DBLP:conf/acl/VeysehLDN20,DBLP:conf/acl/LouLDZC20,DBLP:conf/acl/LiuCX22,DBLP:conf/emnlp/ZhaoJ0GC22} identifies event triggers and their types; 6) \textbf{Event Argument Extraction(EAE)}~\cite{DBLP:conf/emnlp/ZhangSH22,DBLP:conf/emnlp/DuJ22,DBLP:conf/acl/MaW0LCWS22} distinguishes arguments and categorizes their roles with respect to the targe event; and 7) \textbf{Event Extraction(EE)}~\cite{DBLP:conf/emnlp/WaddenWLH19,DBLP:conf/emnlp/DuC20,DBLP:conf/acl/LiuHSW22,DBLP:conf/naacl/HsuHBMNCP22} performs event detection and argument
extraction jointly. Note that although some of these tasks are subsets of others, every task needs LLMs' unique ability to perform well. It is worth to explore the performances on these fine-grained IE tasks.

\subsection{Standard-IE Setting and OpenIE Setting}

To comprehensively evaluate the overall performance of ChatGPT on IE tasks, we ask ChatGPT to generate the responses from the \textbf{Standard-IE} setting and the \textbf{OpenIE} setting. The Standard-IE setting is commonly used in previous works, which uses the task-specific dataset with supervised learning paradigm to fine-tune a model. For ChatGPT, as we can not directly fine-tune the parameters, we evaluate the ChatGPT's ability to select the most appropriate answer from a set of candidate labels instead. Specifically, this setting is based on an instruction that includes the task description, the input text, the prompt, and the label set. Where the task description describes the specific IE task, the prompt involves the utterances that guide the ChatGPT outputs the required keys (which will be introduced in \S~\ref{sec:keys}),  and the label set contains all candidate labels based on each dataset. The \textbf{OpenIE} setting is a more advanced and challenging scenario than \textbf{Standard-IE} setting. In this setting, we do not provide any candidate labels to ChatGPT and rely solely on its ability to comprehend the task description, the prompt, and input text to generate predictions. Our goal is to assess the ChatGPT's ability to produce reasonable factual knowledge.

\begin{table*}[]
\centering
\setlength{\tabcolsep}{0.6mm}
\begin{tabular}{cl}
\toprule[1.5pt]
\multicolumn{1}{c|}{\textbf{Keys}}                    & \multicolumn{1}{c}{\textbf{Explanation}}                                 \\ \hline
\multicolumn{2}{l}{\textit{\textbf{Performance}}}                                                                                \\ \hline
\multicolumn{1}{c|}{\texttt{Open}}                           & Directly ask ChatGPT to predict the class without the label set.                  \\
\multicolumn{1}{c|}{\texttt{Standard}}                         & ChatGPT’s most likely correct class with a given label set. \\
\multicolumn{1}{c|}{\texttt{Top3}}                             & The three most likely classes of the given label set from ChatGPT.                 \\
\multicolumn{1}{c|}{\texttt{Top5}}                             & The five most likely classes of the given label set from ChatGPT.                  \\
\multicolumn{1}{c|}{\texttt{ifOpen\_Correct(Manual)}}             & Manually annotate whether the "\texttt{Open}" is reasonable.             \\ \hline
\multicolumn{1}{l|}{\textit{\textbf{Explainability}}} &                                                                          \\ \hline
\multicolumn{1}{c|}{\texttt{Reason\_Open}}                      & The reason why ChatGPT chooses the class in "\texttt{Open}".                   \\
\multicolumn{1}{c|}{\texttt{Reason\_Standard}}                      & The reason why ChatGPT chooses the class in "\texttt{Standard}".                 \\ 
\multicolumn{1}{c|}{\texttt{ifR\_Open}}                         & Does ChatGPT think that "\texttt{Reason\_Open}" is reasonable?            \\
\multicolumn{1}{c|}{\texttt{ifR\_Standard}}                         & Does ChatGPT think that "\texttt{Reason\_Standard}" is reasonable?          \\ 
\multicolumn{1}{c|}{\texttt{ifR\_Open(Manual)}}                & Manually annotate whether the "\texttt{Reason\_Open}" is reasonable.                    \\
\multicolumn{1}{c|}{\texttt{ifR\_Standard(Manual)}}                 & Manually annotate whether the "\texttt{Reason\_Standard}" is reasonable.             \\      
\hline
\multicolumn{1}{l}{\textit{\textbf{Calibration}}}     &                                                                          \\ \hline
\multicolumn{1}{c|}{\texttt{Confidence\_Open}}                          & The confidence of ChatGPT in predicting "\texttt{Open}".                                       \\
\multicolumn{1}{c|}{\texttt{Confidence\_Standard}}                          & The confidence of ChatGPT in predicting "\texttt{Standard}".                                     \\ \hline
\multicolumn{1}{l}{\textit{\textbf{Faithfulness}}}    &                                                                          \\ \hline
\multicolumn{1}{c|}{\texttt{FicR\_Open(Manual)}}           & Manually annotate whether the "\texttt{Reason\_Open}" is fictitious.                    \\
\multicolumn{1}{c|}{\texttt{FicR\_Standard(Manual)}}           & Manually annotate whether the "\texttt{Reason\_Standard}" is fictitious.                    \\ \bottomrule[1.5pt]
\end{tabular}
\caption{We gather 15 keys in this research, consisting of 10 keys automatically generated by ChatGPT and 5 keys that required manual annotation (denoted as \texttt{Manual}). These keys provide insight into ChatGPT's ability in four dimensions, namely: 1) performance, 2) explainability, 3) calibration, and 4) faithfulness.}
\label{table:key}
\end{table*}

\subsection{Collecting Keys From ChatGPT And Human Annotation}\label{sec:keys}

In this subsection, we first describe 15 keys that are collected from the ChatGPT and domain experts. In Table~\ref{table:key}, we show 10 keys that are extracted from ChatGPT and 5 keys that involves human involvements. These keys could systemically assess ChatGPT's ability from the following four aspects: 

\textbf{Performance.} One important aspect of our research is to comprehensively evaluate the overall performance of ChatGPT on various tasks and compare it with other popular models. By examining its performance from different aspects, we seek to provide a detailed understanding of ChatGPT's capability on the downstream IE tasks. 
    
\textbf{Explainability.} The explainability of ChatGPT is crucial for its application in real-world scenarios~\cite{DBLP:conf/acl/RajaniMXS19,DBLP:conf/acl/AghajanyanGZ20,DBLP:journals/csur/ZiniA23}. In our study, we will measure both the \textit{self-check} and \textit{human-check} explainability of ChatGPT, with a focus on its ability to provide useful and accurate explanations of its reasoning process for humans. Specially, we ask ChatGPT to provide reasons for its predictions (\texttt{Reason\_Open} and \texttt{Reason\_Standard}), and whether ChatGPT approves its explanations (\texttt{ifR\_Open} and \texttt{ifR\_Standard}). Additionally, we also manually evaluate the acceptability of these reasons to humans (\texttt{ifR\_Open(Manual)} and \texttt{ifR\_Standard(Manual)}).
    
\textbf{Calibration.} Measuring the calibration helps to evaluate the predictive uncertainty of a model~\cite{DBLP:conf/icml/GuoPSW17,DBLP:conf/nips/KumarLM19}. A properly calibrated classifier should have predictive scores that accurately reflect the probability of correctness~\cite{DBLP:conf/nips/ThulasidasanCBB19,DBLP:conf/nips/MindererDRHZHTL21}. Given the tendency of modern neural networks to be overconfident in their predictions, we aim to identify potential uncertainties or overconfidence phenomenon of ChatGPT. To evaluate the calibration, ChatGPT is required to provide a confidence score (ranging from 1 to 100) for each prediction it makes (\texttt{Confidence\_Open} and \texttt{Confidence\_Standard}). 
    
\textbf{Faithfulness.} The faithfulness of ChatGPT's explanation is important to ensure its trustworthiness~\cite{DBLP:conf/acl/MaynezNBM20,DBLP:journals/corr/abs-2301-00303}. In evaluating the faithfulness, we have included two keys that assess whether the reasons provided by ChatGPT are faithful to the original input. These keys, \texttt{FicR\_Open(Manual)} and \texttt{FicR\_Standard(Manual)}, require manual annotation by domain experts.

Due to the space limitation, we show an intuitive example in the Appendix~\ref{app:input} to help readers better understand the annotation process.


\section{Performance}\label{sec:main}

\subsection{Setup}
To ensure a comprehensive evaluation of ChatGPT's capabilities, we conduct manual annotation and analysis on a diverse range of IE tasks, including 7 fine-grained tasks spanning 14 datasets. We collected 15 keys for each dataset from both ChatGPT and domain experts (\S~\ref{sec:data_collection}). Only the test sets are annotated, as our aim is to analysis ChatGPT's abilities without any training. For space reasons, the detail of each dataset is shown in the Appendix~\ref{app:data}. Due to the time-consuming nature of obtaining responses from domain experts, we randomly select nearly 3,000 samples in total for our analysis. The number of manually annotated samples for each dataset is reported in the Appendix~\ref{app:data}. As for the outputs from ChatGPT, we use the official API to evaluate the whole test sets.~\footnote{To prevent any historical chat biases, we cleared every conversation after generating each response.} 

Besides, we compare ChatGPT with several popular baselines: 1) \textbf{BERT}~\cite{DBLP:conf/naacl/DevlinCLT19} and \textbf{RoBERTa}~\cite{DBLP:journals/corr/abs-1907-11692}, and 2) \textbf{State-of-the-Art} (SOTA) on the single dataset. Due to the space limitation, the details of state-of-the-art methods are shown in the Appendix~\ref{app:sota}. As for the metric, we use Micro-F1 score for all tasks except RE and EE. For the RE task, we report the named entity recognition F1-score and the relation classification F1-score. As for the EE task, we show the trigger F1-score and argument F1-score.~\footnote{These metrics are all following previous works.}

\subsection{Performance on the Standard-IE Setting}

In this subsection, we report the performances of different models on the Standard-IE setting, as depicted in Table~\ref{tab:main}. It is clear from the table that \textbf{ChatGPT’s performance is not comparable to that of baseline models and SOTA methods in most cases.} This is not surprising given that directly asking ChatGPT for the prediction is more like a zero-shot scenario, whereas the other compared methods are trained on task-specific datasets under a supervised learning paradigm. Another reason may be ChatGPT directly choose an answer from the given label set, and some labels are not easy to understand, thereby negatively impact the performance.


Moreover, our research indicates that \textbf{ChatGPT performs well on relatively simple IE tasks but struggles with more complex and challenging tasks.} For example, the entity typing (ET) task only involves classifying entities into pre-defined types without any further contextual analysis, and ChatGPT excels at this task, demonstrating that the model can generate accurate factual knowledge when the task is simple. However, in complex and challenging IE tasks such as RE, ChatGPT struggles as it requires to first identify the entities that exist in the input and then classify the relationship between them, which is a more challenging task than ET. Despite ChatGPT’s acceptable results on the ET, NER and RC tasks, it still faces challenges with more multifaceted IE tasks like RE and EE, where deeper contextual analysis and reasoning abilities are required. In summary, ChatGPT’s performance varies based on the complexity of the task, and it performs well on straightforward tasks.


Furthermore, the conclusion that ChatGPT performs worse than other models seems inconsistent with previous studies~\cite{DBLP:journals/corr/abs-2302-10205,DBLP:journals/corr/abs-2303-03836}, which suggest that ChatGPT can achieve desirable performance in some IE tasks. One possible explanation for the difference in conclusions is that we report the performance of the entire test set for each task in our study, while prior studies reported on a very small set of test samples drawn at random, which may have substantial variance. Another factor may be that we used a concise and relatively unified prompt to guide ChatGPT, while other research relied on domain-specific prompts or included a large number of label descriptions in their prompts, which needs lots of domain knowledge and thereby limits the ability to generalize across various tasks.

\begin{table*}[]
\setlength{\tabcolsep}{1.0mm}
\begin{tabular}{c|c|c|c|c|c}
\toprule[1.5pt]
\textbf{Task}                                                                                          & \textbf{Dataset}              & \textbf{BERT} & \textbf{RoBERTa} & \textbf{SOTA} & \textbf{ChatGPT}  \\ \hline
\multirow{2}{*}{\textbf{\begin{tabular}[c]{@{}c@{}}Entity\\ Typing(ET)\end{tabular}}}              & \textbf{BBN}           & 80.3          & 79.8             & 82.2~\cite{DBLP:conf/coling/ZuoLJZFL22}          &     85.6       \\ \cline{2-2}
                                                                                                   & \textbf{OntoNotes 5.0} & 69.1          & 68.8             & 72.1~\cite{DBLP:conf/coling/ZuoLJZFL22}          &  73.4                      \\ \hline
\multirow{2}{*}{\textbf{\begin{tabular}[c]{@{}c@{}}Named Entity\\ Recognition(NER)\end{tabular}}}  & \textbf{CoNLL2003}     & 92.8          & 92.4             & 94.6~\cite{DBLP:conf/acl/WangJBWHHT20a}          & 67.2             \\ \cline{2-2}
                                                                                                   & \textbf{OntoNotes 5.0} & 89.2          & 90.9             & 91.9~\cite{DBLP:conf/acl/YeL0S22}          & 51.1            \\ \hline
\multirow{2}{*}{\textbf{\begin{tabular}[c]{@{}c@{}}Relation\\ Classification(RC)\end{tabular}}}    & \textbf{TACRED}        & 72.7      & 74.6             & 75.6~\cite{DBLP:journals/corr/abs-2212-14270}          & 20.3             \\ \cline{2-2}
                                                                                                   & \textbf{SemEval2010}   & 89.1          & 89.8             & 91.3~\cite{DBLP:journals/kbs/ZhaoXCLG21}          & 42.5  \\ \hline
\multirow{2}{*}{\textbf{\begin{tabular}[c]{@{}c@{}}Relation\\ Extraction(RE)\end{tabular}}}        & \textbf{ACE05-R}       & 87.5 | 63.7     & 88.2 | 65.1        &    91.1 | 73.0~\cite{DBLP:conf/acl/YeL0S22}    & 40.5 | 4.5        \\ \cline{2-2}
                                                                                                   & \textbf{SciERC}        & 65.4 | 43.0     & 63.6 | 42.0        &  69.9 | 53.2~\cite{DBLP:conf/acl/YeL0S22}    & 25.9 | 5.5       \\ \hline
\multirow{2}{*}{\textbf{\begin{tabular}[c]{@{}c@{}}Event\\ Detection(ED)\end{tabular}}}            & \textbf{ACE05-E}       & 71.8          &     72.9     & 75.8~\cite{DBLP:conf/acl/LiuCX22}          & 17.1         \\ \cline{2-2}
                                                                                                   & \textbf{ACE05-E+}      & 72.4          & 72.1          & 72.8~\cite{DBLP:conf/acl/LinJHW20}          & 15.5          \\ \hline
\multirow{2}{*}{\textbf{\begin{tabular}[c]{@{}c@{}}Event Argument\\ Extraction(EAE)\end{tabular}}} & \textbf{ACE05-E}       & 65.3          &    68.0       & 73.5~\cite{DBLP:conf/naacl/HsuHBMNCP22}          & 28.9           \\ \cline{2-2}
                                                                                                   & \textbf{ACE05-E+}      & 64.0          &  66.5         & 73.0~\cite{DBLP:conf/naacl/HsuHBMNCP22}          & 30.9   \\ \hline
\multirow{2}{*}{\textbf{\begin{tabular}[c]{@{}c@{}}Event\\ Extraction(EE)\end{tabular}}}           & \textbf{ACE05-E}       & 71.8 | 51.0     &   72.9 | 51.9       & 74.7 | 56.8~\cite{DBLP:conf/acl/LinJHW20}     & 17.0 | 7.3         \\ \cline{2-2}
                                                                                                   & \textbf{ACE05-E+}      & 72.4 | 52.7     &   72.1 | 53.4      & 71.7 | 56.8~\cite{DBLP:conf/naacl/HsuHBMNCP22}     & 16.6 | 7.8      \\\bottomrule[1.5pt]
\end{tabular}
\caption{The performances of ChatGPT and several baseline models on 14 IE datasets on the Standard-IE setting. We report the performance on the whole test set. All results are directly cited from public papers or re-implemented using official open-source code.}
\label{tab:main}
\end{table*}

\begin{table}[]
\centering
\setlength{\tabcolsep}{2.0mm}
\begin{tabular}{c|c|c}
\toprule[1.5pt]
           & \textbf{Standard-IE} & \textbf{OpenIE} \\ \hline
\textbf{BBN}(\textit{ET})     & 86.8\%                      & 97.2\%                    \\
\textbf{CoNLL}(\textit{NER})  &  69.0\%                   & 93.3\%                    \\
\textbf{SemEval2010}(\textit{RC}) & 43.3\%                  & 84.3\%                    \\
\textbf{ACE05-R}(\textit{RE}) &   14.9\%                    & 23.9\%                    \\
\textbf{ACE05-E}(\textit{ED})  & 12.4\%                      & 42.6\%                    \\
\textbf{ACE05-E}(\textit{EAE}) & 17.3\%                      & 65.3\%                    \\
\textbf{ACE05-E}(\textit{EE})  & 4.9\%                       & 28.8\%                    \\ \bottomrule[1.5pt]
\end{tabular}
\caption{The accuracy of Standard-IE setting and OpenIE setting on the sampled test set. Our results show that ChatGPT could generate reasonable outputs on the OpenIE setting.}\label{tab:open}
\end{table}

\subsection{Performance on the OpenIE Setting} 

In this subsection, we report both the accuracy of Standard-IE setting and OpenIE setting on the sampled dataset.~\footnote{We randomly selected around 200 samples for each dataset.} For the Standard-IE setting, we provide the pre-defined label set and ask the ChatGPT to choose an answer for a given input, and the accuracy was calculated by matching the predictions to the ground truth labels. On the other hand, the OpenIE setting refers to asking ChatGPT to make predictions without the pre-defined label set (\texttt{Open} in \S~\ref{sec:keys}). Three domain experts evaluate these predictions and vote on whether they were reasonable in light of the input and background knowledge, named as \texttt{ifOpen\_Correct} in \S~\ref{sec:keys}. Our main goal is to determine if ChatGPT could produce logical and reasonable predictions without given the pre-defined label set, so we do not require the prediction to match with the ground truth.

The results presented in Table~\ref{tab:open} indicate that \textbf{ChatGPT’s performance is somewhat inspiring under the OpenIE setting}. For example, more than 84\% of the predictions are considered reasonable by the domain experts in ET, NER, and RC tasks. However, the performance is relatively poorer for more challenging tasks, such as RE and EE. Overall, compared with Standard-IE setting, ChatGPT's performance on the OpenIE setting is exciting. Our findings suggest that under the OpenIE setting, ChatGPT could generate reliable factual knowledge and reasonable output.

\begin{table*}[h]
\centering
\setlength{\tabcolsep}{2.0mm}
\begin{tabular}{c|ccc|ccc}
\toprule[1.5pt]
\multirow{2}{*}{}    & \multicolumn{3}{c|}{\textbf{Stardand Setting}}                          & \multicolumn{3}{c}{\textbf{OpenIE Setting}}                             \\ \cline{2-7} 
                     & \textbf{Self-check} & \textbf{Human-check} & \textbf{Overlap} & \textbf{Self-check} & \textbf{Human-check} & \textbf{Overlap} \\ \hline
\textbf{BBN} (\textit{ET})        & 100.0\%                & 99.2\%                   & 99.2\%              & 100.0\%                & 99.5\%                   & 99.5\%              \\
\textbf{CoNLL} (\textit{NER})   & 100.0\%                & 99.3\%                   & 99.3\%              & 100.0\%                 & 99.7\%                   & 99.7\%              \\
\textbf{SemEval} (\textit{RC}) & 100.0\%                & 100.0\%                  & 100.0\%             & 100.0\%                & 99.7\%                   & 99.7\%              \\
\textbf{ACE05-R} (\textit{RE})  & 100.0\%                & 90.0\%                   & 90.0\%              & 100.0\%                & 100.0\%                  & 100.0\%             \\
\textbf{ACE05-E} (\textit{ED})  & 100.0\%                & 96.3\%                   & 96.3\%              & 100.0\%                & 90.2\%                   & 90.2\%              \\
\textbf{ACE05-E} (\textit{EAE})   & 100.0\%                & 74.1\%                   & 74.1\%              & 100.0\%                & 90.4\%                   & 90.4\%              \\
\textbf{ACE05-E} (\textit{EE})  & 100.0\%                & 47.1\%                   & 47.1\%              & 94.0\%                 & 78.0\%                   & 74.0\%              \\ \bottomrule[1.5pt]
\end{tabular}
\caption{The explainability of ChatGPT measured on the sampled test set. We report the ratio of samples with reasonable reasons discriminated by ChatGPT (\textit{self-check}) and domain experts (\textit{human-check}) under different settings. Besides, we also compute the overlap ration for both of them. These results indicate that in most cases, ChatGPT exhibits strong explainability for its prediction.}\label{table:reason}
\end{table*}

\subsection{The \textit{top-k} Recall Analysis}

\begin{table}[h]
\centering
\setlength{\tabcolsep}{1.0mm}
\begin{tabular}{c|c|c|c}
\toprule[1.5pt]
\textbf{}            & \textit{\textbf{top-1}} & \textit{\textbf{top-3}} & \textit{\textbf{top-5}} \\ \hline
\textbf{BBN}         & 85.6\%                   & 92.7\%                   & 94.9\% (+9.3\%)                  \\
\textbf{SemEval2010} & 42.5\%                   & 62.1\%                   & 76.0\% (+33.5\%)                  \\\bottomrule[1.5pt]
\end{tabular}
\caption{The \textit{top-k} recall analysis on the whole test set. We report two datasets due to the space limitation, other datasets show similar observation. The results show that ChatGPT could server as a good advisor.}\label{tab:recall}
\end{table}

While generating the most likely prediction may be unsatisfactory on the Standard-IE setting, we seek to investigate whether ChatGPT could be a useful advisor. Therefore, we examine the recall of its \textit{top-k} predictions, with \textit{k} = \textit{1}, \textit{3}, or \textit{5}. As shown in Table~\ref{tab:recall}, the results indicate that compared with the \textit{top-1} recall, the \textit{top-3} recall increases significantly, e.g., the improvement is 19.6\% on SemEval2010. Moreover, the \textit{top-5} recall reaches an impressive 94.9\% on BBN and 76.0\% on SemEval2010, demonstrating a favorable outcome. Our findings suggest that \textbf{ChatGPT is a competent answer candidate generator for a given task under the Standard-IE setting}, which could help users select the most probable prediction from the \textit{top-5} predictions.



\section{Explainability, Calibration and Faithfulness}\label{sec:analysis}
While ChatGPT’s performance in evaluations is noteworthy, it is equally important to evaluate its ability from diverse dimensions that could offer important insights for future research directions. In this section, we analyze several relevant factors, including explainability, calibration, and faithfulness, to comprehensively evaluate ChatGPT’s abilities. Overall, our findings suggest that ChatGPT can provide high-quality and reliable explanations for its predictions, but it tends to display overconfidence in most cases, leading to low calibration. Additionally, ChatGPT displays high faithfulness to the original text, making it an reliable tool for users.

\subsection{Explainability}

Explainability is a critical requirement for LLMs, as it allows users to understand how the model arrives at its predictions~\cite{DBLP:journals/corr/abs-2302-12813}. In this study, we investigate whether ChatGPT could provide a reasonable explanation for its output. To be specific, we request ChatGPT to provide reasons for its predictions in the Standard-IE and OpenIE settings. The corresponding keys are denoted as \texttt{Reason\_Standard} and \texttt{Reason\_Open}, as explained in \S~\ref{sec:keys}. These reasons are then evaluated for their reasonableness by both ChatGPT and three domain experts, with the resulting evaluations referred to as \textit{self-check} and \textit{human-check}, respectively. We only consider the samples with correct predictions in the Standard-IE setting to ensure a robust evaluation of ChatGPT’s explainability ability.~\footnote{We randomly select around 200 samples from each dataset for human annotation.} This is because evaluating the reasons provided by ChatGPT for incorrect predictions is less valuable. 

The ratio of samples with reasonable explanations (termed as \textit{reasonable score}) is summarized in Table~\ref{table:reason}, from which we can derive the following conclusions. Firstly, both \textbf{ChatGPT and domain experts highly approve of the reasons given by ChatGPT}, with the majority of datasets achieving a \textit{reasonable score} of over 90\% in the Standard-IE and OpenIE settings. The above results demonstrate that ChatGPT gives very high-quality explanation for its prediction. Secondly, we observe that \textbf{ChatGPT displays a high level of confidence in the reasons provided for its predictions when compared with human evaluation}. In fact, ChatGPT achieves nearly a 100\% \textit{reasonable score} among almost all datasets. This suggests that ChatGPT is very confident in its ability to provide reasonable explanations. Thirdly, we find that \textbf{when ChatGPT provides a reasonable explanation for a prediction, there is a high level of agreement between ChatGPT and human evaluations}. This suggests that ChatGPT may have a similar understanding of explanations as humans. Overall, our findings suggest that ChatGPT is capable of providing high-quality and reliable explanations for its predictions. This is a crucial step towards developing trustworthy and reliable LLMs.

\begin{table*}[]
\centering
\setlength{\tabcolsep}{3.0mm}
\begin{tabular}{c|ccc|ccc}
\toprule[1.5pt]
                      & \multicolumn{3}{c|}{\textbf{Correct Confidence}}                        & \multicolumn{3}{c}{\textbf{Incorrect Confidence}}                           \\ \cline{2-7} 
                      & \multicolumn{1}{c}{\textbf{BERT}} & \textbf{RoBERTa} & \textbf{ChatGPT} & \textbf{BERT} & \multicolumn{1}{c}{\textbf{RoBERTa}} & \textbf{ChatGPT} \\ \hline
\textbf{BBN}\textit{(ET)}      &           0.971               &    0.968        & 0.888            &   0.904      &    0.885                             & 0.828            \\
\textbf{CoNLL}\textit{(NER)}   &        0.990                &     0.991       & 0.864            &   0.866     &   0.886                             & 0.785            \\
\textbf{SemEval}\textit{(RC)}  &    0.983                        &  0.989            & 0.868            &  0.871         &  0.852                                  & 0.839            \\
\textbf{ACE05-R}\textit{(RE)}  &   0.995                        &    0.991          & 0.760            &  0.883        &  0.810                                & 0.764            \\
\textbf{ACE05-E}\textit{(ED)}  &    0.882                          &  0.944          & 0.852            &  0.770         &  0.871                                & 0.737            \\
\textbf{ACE05-E}\textit{(EAE)} &      0.762                       &  0.785          & 0.956            & 0.525         &  0.555                               & 0.910            \\
\textbf{ACE05-E}\textit{(EE)}  &     0.763                        &  0.782           & 0.845            &  0.612        &  0.628                                & 0.764            \\ \bottomrule[1.5pt]
\end{tabular}
\caption{The prediction confidence of various models on the whole test set. We show both the \textit{correct confidence} and \textit{incorrect confidence} based on various methods. We find that ChatGPT is overconfidence for its prediction in most cases.}\label{tab:confidence}
\end{table*}

\subsection{Calibration}

In this subsection, we first investigate the level of confidence for both the correct and incorrect samples. Confidence is typically described in terms of a probability value, indicating the likelihood of belonging to a specific category. To obtain prediction probabilities from ChatGPT, we ask it to output the probability (\texttt{Confidence\_Standard} and \texttt{Confidence\_Open}), as discussed in \S~\ref{sec:keys}.  Our aim is to investigate whether ChatGPT can provide a reasonable prediction confidence scores for its predictions, thus reducing the risk of misinterpretation. In Table~\ref{tab:confidence}, we present the confidence scores of correct and incorrect predictions from different models, referred to \textit{correct confidence} and \textit{incorrect confidence}, respectively. Our observations reveal that all the models exhibit high confidence levels in their predictions, this is consistent with previous research on large models~\cite{DBLP:conf/icml/GuoPSW17}. \textbf{Although ChatGPT performs worse than its BERT-based counterparts in Standard-IE setting, it displays overconfidence in both correct and incorrect predictions.} Consequently, this overconfidence could lead to misguidance of users. Furthermore, we note a significant confidence gap between correct and incorrect predictions, indicating the need for careful evaluation when ChatGPT's prediction has relatively low confidence.

\begin{table}[h]
\centering
\setlength{\tabcolsep}{0.8mm}
\begin{tabular}{c|ccc}
\toprule[1.5pt]
                      & \multicolumn{1}{c}{\textbf{BERT}} & \textbf{RoBERTa} & \textbf{ChatGPT} \\ \hline
\textbf{BBN}\textit{(ET)}      &        0.012                 &    0.012          & 0.026            \\
\textbf{CoNLL}\textit{(NER)}   &        0.052                      &    0.044          & 0.204            \\
\textbf{SemEval}\textit{(RC)}  &       0.023                      &   0.031          & 0.460            \\
\textbf{ACE05-R}\textit{(RE)}  &       0.020                     &    0.014       & 0.745            \\
\textbf{ACE05-E}\textit{(ED)}  &      0.161                       &   0.226          & 0.656            \\
\textbf{ACE05-E}\textit{(EAE)} &      0.154                       &   0.168          & 0.699            \\
\textbf{ACE05-E}\textit{(EE)}  &      0.211                        &   0.288           & 0.699            \\ \bottomrule[1.5pt]
\end{tabular}
\caption{The expected calibration error (ECE) is used to measure the calibration of a given model, and the lower, the better. Results are calculated on the whole test set.}\label{tab:calibration}
\end{table}



We then focus on calibration, a critical property of LLMs as it could estimate the predictive uncertainty for the secure application of LLMs. A well-calibrated model not only produces accurate predictions but also provides reliable and informative uncertainty estimates, necessary for sound decision-making. In this research, we evaluate the calibration using the Expected Calibration Error (ECE) metric which measures the deviation between predicted confidence and accuracy.~\footnote{We set the bin size to 50, dividing the prediction probabilities into 50 equally spaced bins for analysis.} The results are shown in Table~\ref{tab:calibration}, and from that we can observe that ChatGPT shows much poorer calibration compared to BERT-based methods, which indicates that \textbf{ChatGPT tends to produce confidences that do not represent true probabilities easily}. Furthermore, although ChatGPT displays low ECE in tasks such as ET and NER, miscalibration phenomenon dominates most cases. These findings suggest that ChatGPT needs improvement in terms of calibration, especially for IE tasks.

\subsection{Faithfulness}

\begin{table}[h]
\centering
\setlength{\tabcolsep}{3.0mm}
\begin{tabular}{c|c|c}
\toprule[1.5pt]
                      & \textbf{Stardand-IE} & \textbf{OpenIE} \\ \hline
\textbf{BBN}(\textit{ET})      & 98.3\%                    & 99.3\%                  \\
\textbf{CoNLL}(\textit{NER})   & 100.0\%                    & 98.7\%                  \\
\textbf{SemEval}(\textit{RC})  & 100.0\%                   & 99.1\%                  \\
\textbf{ACE05-R}(\textit{RE})  & 90.0\%                    & 93.8\%                  \\
\textbf{ACE05-E}(\textit{ED})  & 100.0\%                   & 100.0\%                 \\
\textbf{ACE05-E}(\textit{EAE}) & 100.0\%                   & 96.5\%                  \\
\textbf{ACE05-E}(\textit{EE})  & 100.0\%                   & 97.0\%                  \\ \bottomrule[1.5pt]
\end{tabular}
\caption{The evaluation of faithfulness for ChatGPT. Faithfulness refers to whether ChatGPT’s explanation align with the original text. Experimental results show that ChatGPT’s explanation maintains a very high degree of faithfulness to the original text and provide nearly no false explanation.}\label{tab:faithfulness}
\end{table}

Recent works show that ChatGPT may provide false information to users, potentially affecting their decision-making~\cite{DBLP:journals/corr/abs-2302-07736}. Therefore, assessing the faithfulness of the ChatGPT model to the original text is a crucial measurement in developing a trustworthy information extraction model. Our study uses faithfulness as a metric to evaluate the ChatGPT model, specifically referring to if the explanation provided by ChatGPT aligns with the original text when its predictionis correct, as the original text is the most important source for extraction information. There are two keys we collect by domain experts, namely \texttt{FicR\_Standard(Manual)} and \texttt{FicR\_Open(Manual)}, as we mentioned in \S~\ref{sec:keys}. Our results are shown in Table~\ref{tab:faithfulness}, which indicate a high degree of faithfulness between ChatGPT’s explanations and the original text with rare false explanations, i.e., with over 95\% of samples considered faithful in nearly all datasets under different settings. We can conclude that\textbf{ ChatGPT’s decision-making process primarily relies on the input of the original text}, leading to the majority of its explanations being regarded as truthful and reliable.


\section{Conclusion} 

In this paper, we propose to systematically analysis the ChatGPT’s performance, explainability, calibration, and faithfulness. To be specific, based on 7 fine-grained information extraction tasks among 14 datasets, we collect 15 keys identified by either ChatGPT or domain experts for our research. Our findings reveal that ChatGPT’s performance in Standard-IE settings is not as good as BERT-based models in most cases. However, we found that ChatGPT achieved excellent accuracy scores in the OpenIE setting, as evaluated by human annotators. Furthermore, ChatGPT could provide high-quality and trustworthy explanations for its predictions. One of the key issues that we identified is its tendency towards overconfidence, resulting in low calibration. Furthermore, our analysis also showed that ChatGPT exhibits a high level of faithfulness to the original text, indicating that its predictions are grounded in the input text. Given these findings, we hope that our research could inspire more research on using ChatGPT for information extraction.

\bibliography{anthology,custom}
\bibliographystyle{acl_natbib}

\appendix
\clearpage
\section{Appendix}

\subsection{Dataset}\label{app:data}

We report the dataset used in each task in this subsection. For each task, we use two commonly used datasets for the evaluation. Table~\ref{tab:dataset} shows the detailed statistical information. 

Besides, we also report the number of manually annotated samples for each dataset, denoted as \textbf{\#Ann.}, as shown in Table~\ref{table:sample}. 

\begin{table}[H]
\centering
\setlength{\tabcolsep}{4mm}
\begin{tabular}{c|c|c}
\toprule[1.5pt]
\textbf{Task} & \textbf{DataSet} & \textbf{\#Ann.} \\ \hline
\textbf{ED}   & BBN              & 385             \\
\textbf{NER}  & CoNNL            & 300             \\
\textbf{RC}   & SemEval          & 400             \\
\textbf{RE}   & ACE05-R          & 66              \\
\textbf{ED}   & ACE05-E          & 218              \\
\textbf{EAE}  & ACE05-E          & 313              \\
\textbf{EE}   & ACE05-E          & 552       \\\bottomrule[1.5pt]    
\end{tabular}
\caption{The number of manually annotated samples for each dataset.}\label{table:sample}
\end{table}

\begin{table*}[h]
\centering
\setlength{\tabcolsep}{4mm}
\begin{tabular}{c|c|c|c}
\toprule[1.5pt]
\textbf{Task}                                                                                       & \textbf{Dataset}     & \textbf{\#class} & \textbf{\#test} \\ \hline
\multirow{2}{*}{\textbf{\begin{tabular}[c]{@{}c@{}}Entity \\ Typing(ET)\end{tabular}}}              & \textbf{BBN}~\cite{weischedel2005bbn}         & 17               & 23542          \\
                                                                                                    & \textbf{OntoNotes}~\cite{DBLP:journals/corr/GillickLGKH14}   & 41               & 13393          \\ \hline
\multirow{2}{*}{\textbf{\begin{tabular}[c]{@{}c@{}}Named Entity\\ Recognition(NER)\end{tabular}}}   & \textbf{CoNLL 2003}~\cite{DBLP:conf/conll/SangM03}  & 4                & 3453          \\
                                                                                                    & \textbf{OntoNotes}~\footnote{https://catalog.ldc.upenn.edu/LDC2013T19}   & 18               & 8233          \\ \hline
\multirow{2}{*}{\textbf{\begin{tabular}[c]{@{}c@{}}Relation\\ Classification(RC)\end{tabular}}}     & \textbf{TACRED}~\cite{DBLP:conf/emnlp/ZhangZCAM17}      & 42               & 15517           \\
                                                                                                    & \textbf{SemEval2010}~\cite{DBLP:conf/semeval/HendrickxKKNSPP10} & 10               & 2717           \\ \hline
\multirow{2}{*}{\textbf{\begin{tabular}[c]{@{}c@{}}Relation\\ Extraction(RE)\end{tabular}}}         & \textbf{ACE05-R}~\footnote{https://catalog.ldc.upenn.edu/LDC2006T06}     &   7/6        & 2050            \\
                                                                                                    & \textbf{SciERC}~\cite{DBLP:conf/emnlp/LuanHOH18}      &    6/7    & 551           \\ \hline
\multirow{2}{*}{\textbf{\begin{tabular}[c]{@{}c@{}}Event\\ Detection(ED)\end{tabular}}}             & \textbf{ACE05-E}~\cite{DBLP:conf/emnlp/WaddenWLH19}     & 33            & 832             \\
                                                                                                    & \textbf{ACE05-E$^{+}$}~\cite{DBLP:conf/acl/LinJHW20}    & 33            & 676            \\ \hline
\multirow{2}{*}{\textbf{\begin{tabular}[c]{@{}c@{}}Event Argument \\ Extraction(EAE)\end{tabular}}} & \textbf{ACE05-E}~\cite{DBLP:conf/emnlp/WaddenWLH19}     & 22/33            & 403             \\
                                                                                                    & \textbf{ACE05-E$^{+}$}~\cite{DBLP:conf/acl/LinJHW20}    & 22/33            & 424             \\ \hline
\multirow{2}{*}{\textbf{\begin{tabular}[c]{@{}c@{}}Event \\ Extraction(EE)\end{tabular}}}           & \textbf{ACE05-E}~\cite{DBLP:conf/emnlp/WaddenWLH19}     & 22/33            & 832            \\
                                                                                                    & \textbf{ACE05-E$^{+}$}~\cite{DBLP:conf/acl/LinJHW20}    & 22/33            & 676            \\ \bottomrule[1.5pt]  
\end{tabular}
\caption{The table presents several key statistical characteristics of the datasets used in our research, including 14 datasets that belonging to 7 different IE tasks.}
\label{tab:dataset}
\end{table*}

\subsection{The State-of-the-Art Methods on Single Dataset}\label{app:sota}

In this section, we introduce the state-of-the-art method on each dataset:

\textbf{Entity Typing (ET):} \citet{DBLP:conf/coling/ZuoLJZFL22} proposed a type-enriched hierarchical contrastive strategy for entity typing task, named \textbf{PICOT}. \textbf{PICOT} models differences between hierarchical types to distinguish similar types at different levels of granularity. It also embeds type information into entity contexts and employ a constrained contrastive strategy on the hierarchical structure. This method achieves SOTA results on the \textbf{BBN} and \textbf{OntoNotes 5.0} datasets.

\textbf{Named Entity Recognition (NER):} \citet{DBLP:conf/acl/WangJBWHHT20a} proposed a model named \textbf{ACE}, which automates finding better embeddings for structured prediction tasks. It uses a neural architecture search-inspired formulation where a controller updates belief based on a reward. The reward is the accuracy of a task model trained on a task dataset with the concatenated embeddings as input. This method achieves SOTA results on the \textbf{CoNLL2003} dataset. 

\textbf{Relation Classification (RC):} \citet{DBLP:journals/corr/abs-2212-14270} proposed Label Graph Network with \textit{Top-k} Prediction Set (\textbf{KLG}), to effectively utilize the \textit{Top-k} prediction set. \textbf{KLG} builds a label graph for a given sample to review candidate labels in the \textit{Top-k} prediction set and learns the connections between them. It also includes a dynamic k-selection mechanism to learn more powerful and discriminative relation representation. This method sets SOTA results on the \textbf{TACRED} dataset. \citet{DBLP:journals/kbs/ZhaoXCLG21} proposed \textbf{RIFRE}, which models relations and words as nodes on a graph, and iteratively fuses the two types of semantic nodes using message passing. This approach obtains node representations that are better suited for relation extraction tasks. The model then performs relation extraction on the updated node representations. This method sets SOTA results on the \textbf{SemEval2010} dataset. 

\textbf{Relation Extraction (RE):} \citet{DBLP:conf/acl/YeL0S22} proposed \textbf{PL-Marker}, a novel span representation approach that considers the interrelation between span pairs by packing markers in the encoder. To better model entity boundary information, \textbf{PL-Marker} proposes a neighborhood-oriented packing strategy that considers neighbor spans integrally. For more complicated span pair classification tasks, this paper also designs a subject-oriented packing strategy, which packs each subject and its objects to model the interrelation between same-subject span pairs. This method sets SOTA results on \textbf{ACE05-R} and \textbf{SciERC} datasets. It also achieves the best performance on the \textbf{OntoNotes 5.0} datasets of NER task.

\textbf{Event Detection (ED):} \citet{DBLP:conf/acl/LiuCX22} proposed \textbf{SaliencyED}, a novel training mechanism
for ED, which can distinguish between trigger-dependent and context-dependent types, and achieves promising results on \textbf{ACE05-E} dataset. \citet{DBLP:conf/acl/LinJHW20} proposed \textbf{ONEIE} neural framework aims to globally optimize information extraction as a graph from an input sentence, capturing cross-subtask and cross-instance inter-dependencies, and and achieves promising results on \textbf{ACE05-E+} dataset.

\textbf{Event Argument Extraction (EAE):} \citet{DBLP:conf/naacl/HsuHBMNCP22} proposed \textbf{DEGREE}, which formulates event extraction as a conditional generation problem, summarizing events mentioned in a passage into a natural sentence following a predefined pattern, as learned from a prompt. Extracted event predictions are then obtained from the generated sentence using a deterministic algorithm. \textbf{DEGREE} sets the best results on both \textbf{ACE05-E} and \textbf{ACE05-E+} datasets.

\textbf{Event Argument Extraction (EAE):} the SOTA methods are \textbf{ONEIE} for \textbf{ACE05-E}, and \textbf{DEGREE} for \textbf{ACE05-E+}.

\subsection{Exemplar of the Input}\label{app:input}
In this section, we show an input examples for the event detection task to help readers understand our implement, as shown in Table~\ref{app:ed}.

\begin{table*}[h]
\centering
\begin{tabular}{l}
\toprule[1.5pt]
\textit{\textbf{Input of Event Detection (ED)}} \\ \hline
\textbf{Task Description:} Given an input list of words, identify all triggers in the list, and categorize \\ each of them into the predefined set of event types. A trigger is the main word that most clearly \\ expresses the occurrence of an event in the predefined set of event types.  \\ \hline
\textbf{Pre-defined Label Set:} The predefined set of event types includes: {[}Life.Be-Born, Life.Marry, \\ Life.Divorce, Life.Injure, Life.Die, Movement.Transport, Transaction.Transfer-Ownership, \\ Transaction.Transfer-Money, Business.Start-Org, Business.Merge-Org, Business.Declare-\\Bankruptcy, Business.End-Org, Conflict.Attack, Conflict.Demonstrate, Contact.Meet, Contact.\\Phone-Write, Personnel.Start-Position, Personnel.End-Position, Personnel.Nominate, Personnel.\\Elect, Justice.Arrest-Jail, Justice.Release-Parole, Justice.Trial-Hearing, Justice.Charge-Indict, \\Justice.Sue, Justice.Convict, Justice.Sentence, Justice.Fine, Justice.Execute, Justice.Extradite, \\Justice.Acquit, Justice.Appeal, Justice.Pardon{]}. \\ \hline
\textbf{Input and Task Requirement:} Perform ED task for the following input list, and print the output: \\{[}'Putin', 'concluded', 'his', 'two', 'days', 'of', 'talks', 'in', 'Saint', 'Petersburg', 'with', 'Jacques', \\'Chirac', 'of', 'France', 'and', 'German', 'Chancellor', 'Gerhard', 'Schroeder', 'on', 'Saturday', \\'still', 'urging', 'for', 'a', 'central', 'role', 'for', 'the', 'United', 'Nations', 'in', 'a', 'post', '-', \\'war', 'revival', 'of', 'Iraq', '.'{]} The output of ED task should be a list of dictionaries following \\json format. Each dictionary corresponds to the occurrence of an event in the input list and should \\consists of "trigger", "word\_index", "event\_type", "top3\_event\_type", "top5\_event\_type", \\"confidence", "if\_context\_dependent",  "reason" and "if\_reasonable" nine keys. The value of "word\_\\index" key is an integer indicating the index (start from zero) of the "trigger" in the input list. The \\value of "confidence" key is an integer ranging from 0 to 100, indicating how confident you are that \\the "trigger" expresses the "event\_type" event. The value of "if\_context\_dependent" key is either 0 \\(indicating the event semantic is primarily expressed by the trigger rather than contexts) or 1 \\(indicating the event semantic is primarily expressed by contexts rather than the trigger). The value \\of "reason" key is a string describing the reason why the "trigger" expresses the "event\_type", and \\do not use any " mark in this string. The value of "if\_reasonable" key is either 0 (indicating the reason \\given in the "reason" field is not reasonable) or 1 (indicating the reason given in the "reason" field is \\reasonable). Note that your answer should only contain the json string and nothing else.

                \\ \bottomrule[1.5pt]
\end{tabular}
\caption{The input example of event detection task. This example is extracted from ACE05-E, and all the above three parts are jointly imported into ChatGPT.}\label{app:ed}
\end{table*}

\end{document}